# Empirical Evaluation of Approximation Algorithms for Probabilistic Decoding

Irina Rish, Kalev Kask and Rina Dechter*
Department of Information and Computer Science
University of California, Irvine
{irinar, kkask, dechter}@ics.uci.edu

## Abstract

It was recently shown that the problem of decoding messages transmitted through a noisy channel can be formulated as a belief updating task over a probabilistic network [14]. Moreover, it was observed that iterative application of the (linear time) belief propagation algorithm designed for polytrees [15] outperformed state of the art decoding algorithms, even though the corresponding networks may have many cycles.

This paper demonstrates empirically that an approximation algorithm *approx-mpe* for solving the most probable explanation (MPE) problem, developed within the recently proposed mini-bucket elimination framework [4], outperforms iterative belief propagation on classes of coding networks that have bounded *induced width*. Our experiments suggest that approximate MPE decoders can be good competitors to the approximate belief updating decoders.

## 1 Introduction

In this paper we evaluate the quality of a recently proposed *mini-bucket* approximation scheme [8, 6] for probabilistic decoding.

Recently, a class of parameterized *mini-bucket approximation* algorithms for probabilistic networks, based on the *bucket elimination* framework, was proposed in [4]. The approximation scheme uses as a controlling parameter a bound on arity of probabilistic functions (i.e., clique size) created during variable elimination, allowing a trade-off between accuracy and efficiency [8, 6]. The mini-bucket algorithms were presented and analyzed for several tasks, such as belief updating, finding the most probable explanation (MPE) and finding the maximum a posteriori hypothesis (MAP). Encouraging results were obtained on randomly generated *noisy-OR* networks and on the CPCS networks [16]. Clearly, more testing is necessary to determine the regions of applicability of this class of algorithms. In particular, testing on realistic applications is mandatory.

When it was recently shown that the problem of decoding noisy messages can be described as a belief updating task over a probabilistic network [14], we decided to use this problem as our next benchmark. This domain was particularly interesting since the recent advances in probabilistic decoding demonstrated that the (linear time) *belief propagation* algorithm for polytrees [15] yields a nearly optimal decoder if applied iteratively to multiply-connected coding networks, such as *low-density parity-check* codes [13], *low-density generator matrix* codes [3], and, especially, *turbo codes* [2]. This is considered "the most exciting and potentially important development in coding theory in many years" [14]. Initial analysis of iterative belief propagation on networks with loops is presented in [20].

In this paper we compare the mini-bucket MPE approximation algorithm *approx-mpe* to the iterative belief propagation (*IBP*), and to the exact elimination algorithms *elim-mpe*, *elim-map* and *elim-bel* for MPE, MAP and for belief updating, respectively, on several classes of *linear block codes*, such as *Hamming codes*, *randomly* generated block codes, and *structured* low-induced-width block codes. A codeword is transmitted through a channel adding a Gaussian noise to each bit. Given the real-valued *channel output y*, the task of decoding is to find the most likely assignment either to each information bit (bit-wise decoding), or to the whole information sequence (block-wise decoding) [14, 10, 11]. Other variation (not considered here) could be to round $y$ to a 0/1 vector before decoding.

This paper demonstrates empirically that
1. on a class of structured codes having low induced width the mini-bucket approximation *approx-mpe* outperforms *IBP*;
2. on a class of random networks having large induced width and on some Hamming codes *IBP* outperforms

*This work was partially supported by NSF grant IRI-9157636 and by Air Force Office of Scientific Research grant, AFOSR 900136, Rockwell International and Amada of America.

*approx-mpe*;
3. as expected, exact MPE decoding, *elim-mpe*, outperforms approximate decoding. However, on random networks exact MPE decoding was not feasible due to the large *induced width* (as it is known, variable elimination algorithms are exponential in the induced width of the network).
4. The *exact* MPE (maximum-likelihood) (*elim-mpe*) and the *exact* belief update decoding (*elim-bel*) have comparable error for relative low channel noise; for larger noise, belief update decoding gives a slightly smaller (by $\approx 0.1\%$) bit error than the MPE decoding.

## 2 Definitions and notation

**Definition 1:** [graph concepts] A *directed graph* is a pair $G = \{V, E\}$, where $V = \{X_1, ..., X_n\}$ is a set of nodes, or variables, and $E = \{(X_i, X_j) | X_i, X_j \in V\}$ is the set of edges. Given $(X_i, X_j) \in E$, $X_i$ is called a *parent* of $X_j$, and $X_j$ is called a *child* of $X_i$. The set of $X_i$'s parents is denoted $pa(X_i)$, or $pa_i$, while the set of $X_i$'s children is denoted $ch(X_i)$, or $ch_i$. The family of $X_i$ includes $X_i$ and its parents. A directed graph is acyclic if it has no *directed* cycles. A graph is *singly connected* (also called a *polytree*), if its underlying undirected graph has no cycles. Otherwise, it is called *multiply connected*. An *ordered graph* is a pair $(G, o)$ where $o = X_1, ..., X_n$ is an ordering of the nodes in the graph $G$. The *width of a node* in an ordered graph is the number of the node's neighbors that precede it in the ordering. The *width of an ordering o*, denoted $w(o)$, is the maximum width over all nodes. The *induced graph* of $G$ along an ordering $o$ is obtained by connecting the preceding neighbors of each $x_i$, going from $x_N$ to $x_1$. The *induced width of a graph along an ordering o*, $w_o^*$, is the width of the induced graph along $o$. The *induced width of a graph*, $w*$, is the minimal induced width over all its orderings; $w*+1$ is also called *tree-width* [1]. For more information see [9, 7]. The *moral graph* of a directed graph $G$ is the undirected graph obtained by connecting the parents of all the nodes in $G$ and removing the arrows.

**Definition 2:** [belief networks] Let $X = \{X_1, ..., X_n\}$ be a set of random variables over multivalued domains $D_1, ..., D_n$. A *belief network (BN)* is a pair $(G, P)$ where $G$ is a directed acyclic graph on $X$ and $P = \{P(X_i | pa_i) | i = 1, ..., n\}$ is the set of conditional probability matrices associated with each $X_i$. An assignment $(X_1 = x_1, ..., X_n = x_n)$ can be abbreviated as $x = (x_1, ..., x_n)$. The BN represents a joint probability distribution $P(x_1, ..., x_n) = \Pi_{i=1}^{n} P(x_i | x_{pa(X_i)})$, where $x_S$ is the projection of vector $x$ on a subset of variables $S$. An evidence $e$ is an instantiated subset of variables.

Given evidence $e$, the following tasks can be defined over a belief network. 1. *Belief updating*: finding the posterior probability $P(Y|e)$ of a *query* nodes $Y \in X$; 2. *Most probable explanation* (MPE): finding an assignment $x^o = (x^o_1, ..., x^o_n)$, such that $p(x^o) = \max_x P(x|e)$. 3. *Maximum aposteriory hypothesis* (MAP): finding an assignment $a^o = (a^o_1, ..., a^o_k)$ to a subset of variables $A = \{A_1, ..., A_k\}$, called a *hypothesis*, such that $p(a^o) = \max_{\bar{a}_k} \sum_{x_{X-A}} P(x|e)$, where $\bar{a}_k = \{A_1 = a_1, ..., A_k = a_k\}$.

## 3 Noisy channel coding

The purpose of *channel coding* is to provide reliable communication through a noisy channel. A *systematic error-correcting encoding* [14] maps a vector of $K$ *information bits* $u = (u_1, ..., u_K), u_i \in \{0, 1\}$, into an $N$-bit *codeword* $c = (u, x)$, where $N - K$ additional bits $x = (x_1, ..., x_{N-K}), x_j \in \{0, 1\}$ add redundancy to the information source in order to decrease the decoding error. The codeword, called the *channel input*, is transmitted through a noisy channel. A commonly used Additive White Gaussian Noise (AWGN) channel model implies that independent Gaussian noise with variance $\sigma^2$ is added to each transmitted bit, producing the *channel output y*. Given a *real-valued* vector $y$, the *decoding* task is to restore the input information vector $u$ [10, 14, 13]. An alternative approach, not considered here, is to round $y$ to a 0/1 vector before decoding.

The probability of decoding error is measured by the *bit error rate (BER)*, i.e. the percentage of incorrectly decoded bits. The *code rate* $R = K/N$ is the fraction of the information bits to the total number of transmitted bits. The goal is to minimize the decoding error while maintaining a relatively low code rate. Shannon [19] proved that there exists a lower bound, called *Shannon's limit*, to the decoder's performance. Given the noise variance $\sigma^2$ and a fixed code rate $R$, Shannon's limit gives the smallest achievable BER no matter which code is used. Unfortunately, Shannon's proof was nonconstructive, leaving open the problem of finding optimal codes. Moreover, it assumes an optimal maximum-likelihood decoding, which is intractable for high-performance codes that tend to be long [17]. "As far as is known, a code with a low-complexity optimal decoding algorithm cannot achieve high performance" [14]. Therefore, good codes should be combined with efficient approximate decoding algorithms into high-performance *coding schemes*.

Recently, several low-error coding schemes have been proposed, such as turbo-codes [2], low-density parity-check codes [13] and low-density generator matrix codes [3], that achieve a near-Shannon-limit performance. It was shown, that the decoding algorithms used by turbo-codes and some other low-error codes can be viewed as an iterative application of the *belief propagation* algorithm [15], designed for polytrees, to multiply-connected coding networks.

In this paper, we report on experiments with several types of *linear (N,K) block codes*. A linear (N,K) block code can be defined in terms of a *generator matrix* or in terms of a *parity check* matrix [14]. Given a 0/1 *generator matrix* G, and an information vector $u$, the codeword $c = (u, x)$ must satisfy $c = uG$, where summation modulo 2 is assumed. For example, the generator matrix

$$G = \begin{matrix} 1 & 0 & 0 & 0 & 1 & 1 & 0 \\ 0 & 1 & 0 & 0 & 1 & 0 & 1 \\ 0 & 0 & 1 & 0 & 0 & 1 & 1 \\ 0 & 0 & 0 & 1 & 1 & 1 & 1 \end{matrix}$$

defines a (7,4) *Hamming code*. It can be depicted by the belief network in Figure 1, that represents each bit in $u$, $x$, and $y$ vectors as a node. The parent set for each $x_i$ corresponds to non-zero entries in $(K + i)$th column of $G$, and the *deterministic* conditional probability function $P(x_i|pa_i)$ equals 1 iff $x_i = u_{j_1} \oplus ... \oplus u_{j_p}$, where $\oplus$ is addition modulo 2. The bits $x_i$ are also called the *parity-check* bits. Note, that this structure is a special (deterministic) case of *causal independence*, which can be exploited to speed up probabilistic inference [12, 21, 18]. Each output bit $y_j$ has exactly one parent, the corresponding channel input bit. The conditional density function $P(y_j|c_j)$ is a Gaussian (normal) distribution $N(c_j; \sigma)$, where the mean equals the value of the transmitted bit, and $\sigma^2$ is the noise variance.

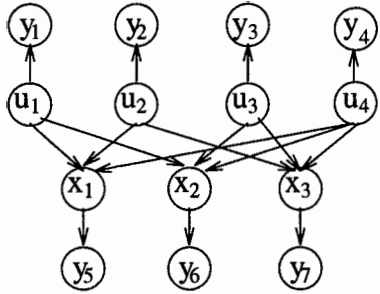

Figure 1: Belief network for a (7,4) Hamming code

Probabilistic decoding can be formulated either as maximum aposteriory *information bit* decoding, or as a maximum aposteriory (maximum-likelihood) *information sequence* (*block*) decoding. Formally, given the observed output $y$, the first task is to find for each $k$

$$u_k^* = \arg\max_{u_k} P(u_k|y), 1 \leq k \leq K,$$

while the second is to find

$$u' = \arg\max_u P(u|y).$$

An alternative formulation of block-wise decoding is to find a most probable explanation (MPE) assignment to all codeword bits

$$(u', x') = \arg\max_{(U,X)} P(U, X|Y).$$

Therefore, the bit-wise decoding requires finding posterior probabilities for each information bit and can be solved by belief updating algorithms, while the block-wise decoding requires algorithms for finding MAP or MPE. In the next section, we discuss exact and approximate algorithms for these tasks.

---

**Algorithm elim-mpe**
**Input:** A belief network $BN = \{P_1, ..., P_n\}$; a variable ordering $o$; an evidence $e = \{(X_i = v_i)| X_i \in Y, Y \subset X\}$.
**Output:** The most probable assignment $X^*$.
1. **Initialize:** Partition $BN$ into $bucket_1, ..., bucket_n$, s.t. $bucket_i$ contains all $P_j$ whose highest variable is $X_i$. Put each observation $(X_i = v_i)$ into $bucket_i$. Let $h_1, h_2, ..., h_j$ be the functions in $bucket_p$, and let $S_1, ..., S_j$ be the subsets of variables on which $h_j$ are defined.
2. **Backward:** For $p \leftarrow n$ downto 1, do
• If $bucket_p$ contains $X_p = v_p$, /* evidence */ assign $v_p$ to $X_p$ in each $h_i$ and put the result into an appropriate bucket.
• Else, compute $h^p$: $h^p = max_{X_p} \Pi_{i=1}^{j} h_i$. Add $h^p$ to the bucket of the largest-index variable in $U_p \leftarrow \bigcup_{i=1}^{j} S_i - \{X_p\}$.
3. **Forward:** Assign optimal values to $X_p$ along the ordering $o$, as follows: $X_p^* = \arg\max_{X_p} h^p$.

Figure 2: Algorithm *elim-mpe*

## 4 Algorithms

In this section we present the algorithms to be evaluated on coding problems. We start with a brief description of the *bucket-elimination* framework, that includes the algorithms *elim-bel*, *elim-mpe*, and *elim-map* for computing exact belief, MPE, and MAP, respectively [4], and the *mini-bucket* approximation *approx-mpe(i)* for MPE [8, 6]. Then we present the *Iterative Belief Propagation (IBP)* algorithm.

### 4.1 Bucket-elimination algorithms

*Bucket elimination* was introduced recently as a unifying variable elimination algorithmic framework for probabilistic and deterministic reasoning [5]. In particular, it was shown that many algorithms for probabilistic inference, such as belief updating, finding the most probable explanation, finding the maximum aposteriori hypothesis, and calculating the maximum expected utility, can be expressed as bucket-elimination algorithms [4].

The input to a bucket-elimination algorithm consists of a knowledge-base theory specified by a collection of functions or relations, (e.g., clauses for propositional satisfiability, constraints, or conditional probability matrices for belief networks). Given a variable ordering, the algorithm partitions the functions into buckets, each associated with a single variable, and process the buckets in reverse order, from last variable to first. Each bucket is processed by some vari-

---

**Algorithm approx-mpe(i)**
**Input:** A belief network $BN = \{P_1, ..., P_n\}$; a variable ordering $o$; an evidence $e = \{(X_i = v_i)| X_i \in Y, Y \subset X\}$.
**Output:** An upper and lower bounds on the most probable assignment.
1. **Initialize:** Partition $BN$ into $bucket_1, ..., bucket_n$, s.t. $bucket_i$ contains all $P_j$ whose highest variable is $X_i$. Put each observation $(X_i = v_i)$ into $bucket_i$.
Let $h_1, h_2, ..., h_j$ be the functions in $bucket_p$, and let $S_1, ..., S_j$ be the subsets of variables on which $h_j$ are defined.
2. **Backward:** For $p \leftarrow n$ downto 1, do
• (evidence) If $bucket_p$ contains $X_p = v_p$, assign $v_p$ to $X_p$ in each $h_i$ and put the result into an appropriate bucket.
• **else,** //for $h_1, h_2, ..., h_j$ in $bucket_p$, do:
Generate an $i$-partitioning, $Q' = \{Q_1, ..., Q_r\}$, of functions $h_1, h_2, ..., h_j$ in $bucket_p$.
For each $Q_l \in Q'$ containing $h_{l_1}, ... h_{l_t}$ do,
Generate function $h^l$, $h^l = max_{X_p} \Pi_{i=1}^{t} h_{l_i}$. Add $h^l$ to the bucket of the largest-index variable in $U_l \leftarrow \bigcup_{i=1}^{j} S_{l_i} - \{X_p\}$.
3. **Forward:** For $p = 1$ to $n$ do: given $x_1, ..., x_{p-1}$, assign to $X_p$ a value $v_p$ that maximizes the product of all the functions in $bucket_p$.

---

Figure 3: algorithm *approx-mpe(i)*

able elimination procedure over the functions in the bucket, and the resulting new function is placed in a lower bucket.

Figure 2 shows *elim-mpe*, the bucket-elimination algorithm for computing MPE [4]. Given a variable ordering, the conditional probability matrices are partitioned into buckets, where each matrix is placed in the bucket of the highest variable it mentions. When processing the bucket of $X_p$, a new function is generated by taking the maximum relative to $X_p$, over the product of functions in that bucket. The resulting function is placed in the appropriate lower bucket. The algorithms *elim-bel* and *elim-map* for belief updating and for finding MAP are derived in a similar way [4].

The complexity of the bucket elimination algorithms is determined by the complexity of processing each bucket (step 2), and is time and space exponential in the number of variables in the bucket. This number equals $w^* + 1$, where $w^*$ is the *induced-width* of the of the network's moral graph along the given variable ordering.

### 4.2 Mini-bucket algorithms

Since elimination becomes intractable when the functions $h_p$ created in the buckets are too large, we proposed in [8] to approximate these functions by a collection of smaller functions. Let $h_1, ..., h_j$ be the functions in the bucket of $X_p$, and let $S_1, ..., S_j$ be the variable subsets on which those functions are defined. When *elim-mpe* processes the bucket of $X_p$, it computes the function $h^p$: $h^p = max_{X_p} \Pi_{i=1}^{j} h_i$. A brute-force approximation method involves migrating the maximization operator inside the multiplication, generating instead a new function $g^p$: $g^p = \Pi_{i=1}^{j} max_{X_p} h_i$. Obviously, $h^p \leq g^p$. Each maximized function will have the arity lower than the arity of $h_i$, and each of these functions can be moved, separately, to a lower bucket. When the algorithm reaches the first variable, it has computed an *upper bound* on the MPE. A maximizing tuple can then be generated by instantiating the variables going from first bucket to last, using the information that is recorded in each bucket. The probability of the tuple is the *lower bound* on the MPE.

This idea was extended to yield a collection of parameterized approximation algorithms by partitioning the bucket into mini-buckets of varying sizes and applying the elimination operator on each mini-bucket rather then on single function. This yields approximations of varying degrees of accuracy and efficiency. Let $Q' = \{Q_1, ..., Q_r\}$ be a partitioning into mini-buckets of the functions $h_1, ..., h_j$ in $X_p$'s bucket. If the mini-bucket $Q_l$ contains the functions $h_{l_1}, ..., h_{l_r}$, the approximation will compute $g^p = \Pi_{l=1}^{r} max_{X_p} \Pi_{l_i} h_{l_i}$. Clearly, coarser partitionings yield the higher accuracy, but also the higher complexity.

Algorithm *approx-mpe(i)* is described in Figure 3. It is parameterized by the bound $i$ on the number of variables allowed in a mini-bucket. A partitioning $Q$ into mini-buckets is an $i$-partitioning if the total number of variables in each mini-bucket does not exceed $i$.

It was shown [8] that algorithm approx-mpe($i$) computes an upper bound to the MPE in time and space $O(exp(i))$ where $i \leq n$.

### 4.3 Iterative belief propagation

*Iterative Belief Propagation* (IBP) computes an approximate belief for every variable in the network. It applies Pearl's belief propagation algorithm [15], developed for singly-connected networks, to a multiply-connected networks, ignoring cycles. Belief is propagated by sending messages between the nodes: for each node $x$, its parents $u_i$ send *causal* support messages $\pi_{u_i,x}$, while its children $y_j$ send *diagnostic* support messages $\lambda_{y_j,x}$. Causal supports from all parents and diagnostic support from all children are combined into vectors $\pi_x$ and $\lambda_x$, respectively.

An *activation schedule* (variable ordering) $A$ specifies the order in which the nodes are processed (activated). After all nodes are processed, the next iteration of belief propagation begins, updating the messages computed during the previous iteration. Algorithm IBP(n) stops after $n$ iterations. Note, that the algorithm is not guaranteed to converge to the correct beliefs on multiply-connected networks.

The algorithm is given in figure 4. It takes a variable ordering as an input. We have used an ordering that

> **Iterative Belief Propagation (IBP)**
> **Input:** A Belief Network $BN = \{P_1, ..., P_n\}$, evidence $e$, an activation schedule $A$, the number of iterations $I$.
> **Output:** Belief in every variable.
>
> 1. Initialize $\lambda$ and $\pi$.
> For evidence node $x_i = j$, set $\lambda_{x_i}(k)$ to 1 for $j = k$ and to 0 for $j \neq k$. For node $x$ having no parents set $\pi_x$ to prior $P(x)$. Otherwise, set $\lambda_x$ and $\pi_x$ to $(1,...,1)$.
>
> 2. **For iterations** = 1 to $I$:
> For each node $x$ along $A$, do:
> /* $\alpha$ is a normalization constant */
> • For each $x = j$, compute $\lambda_{x,u_i}(j) =$
> $\alpha \sum_j \lambda_x(j) \sum_{u_l: l \neq i} P(x = j \mid u_1, ..., u_m) \prod_{l \neq i} \pi_{u_l, x}$
> • For each $x = j$, compute $\pi_{x, y_i}(j) =$
> $\alpha \prod_{k \neq i} \lambda_{y_k, x}(j) \sum_{u_l} P(x = j \mid u_1, ..., u_m) \prod_l \pi_{u_l, x}$.
>
> 3. **Belief update:** For each $x$ along $A$
> • For each $x = j$, compute
> $\lambda_x(j) = \prod_i \lambda_{y_i, x}(j)$, where $y_i$ are $x$'s children.
> • For each $x = k$, compute
> $\pi_x(j) = \sum_{u_1, ..., u_m} P(x = j \mid u_1, ..., u_m) \prod_i \pi_{u_i, x}$,
> where $u_i$ are $x$'s parents.
> • Compute $BEL(x) = \alpha \lambda_x \pi_x$.

Figure 4: Iterative Belief Propagation (IBP) Algorithm

first updates the input variables of the coding network and then updates the parity-check variables. Notice that evidence variables are not updated.

## 5 Experimental Methodology

We experimented with several types of $(N, K)$ linear block codes, which include $(7, 4)$ and $(15, 11)$ Hamming codes, randomly generated codes, and structured codes with low induced width. So far, we used only the code rate $R = 1/2$, i.e. $N = 2K$.

Both random and structured code networks have a fixed number of parents, $P$. In random codes, for each parity-check bit $x_j$, $P$ parents are selected randomly out of $K$ information bits. In structured parity-check codes, each parity bit $x_i$ has $P$ sequential parents $\{u_{(i+j) \bmod K}, 0 \leq j < P\}$.

Our random and structured codes can be represented as four-layer belief networks having $K$ nodes in each layer (see Figure 5). The two inner layers (channel input nodes) correspond to the input information bits $u_i, 0 \leq i < K$, and to the parity-check bits, $x_i, 0 \leq i < K$. The two outer layers represent the channel output $y = (y^u, y^x)$, where $y^u$ and $y^x$ result from transmitting $u$ and $x$, respectively. The input nodes are binary (0/1), while the output nodes are real-valued.

Given K, P, and the channel noise variance $\sigma^2$, a sample coding network is generated as follows. First, the appropriate belief network structure is created. Then, we simulate an input signal assuming uniform random distribution of information bits, and the corresponding values of the parity-check bits are computed. Finally, we generate an assignment to the observed vector $y$ by adding Gaussian noise to each information and parity-check bit. The decoding algorithm takes as an input the coding network and the observed channel output $y$ (a real-valued assignment to the $y_i^u$ and $y_j$ nodes). The task is to recover the original information sequence.

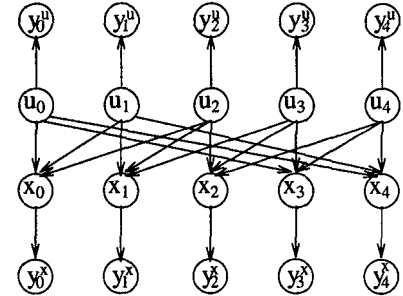

Figure 5: Belief network for structured (10,5) block code with parent set size P=3

We experimented with the following decoding algorithms: iterative belief propagation (IBP), the exact elimination algorithms for belief updating and finding MAP and MPE (elim-bel, elim-map and elim-mpe, respectively), and approx-mpe(i) algorithm. When $i < P$, the number of variables in a mini-bucket is bounded by $max(i, P + 1)$, rather than by $i$. $IBP(I)$ denotes iterative belief propagation with $I$ iterations.

For each decoding algorithm, we plot the observed bit error rate (BER) versus the channel parameter $\sigma$.

In all our experiments the BER of elim-map coincided with that of elim-mpe, and will not be reported separately.

## 6 Results and Discussion

The results on the three classes of coding networks are summarized in Figures 6(a)-(g). For more details, see table 2. On the class of parity-check codes we ran up to 10 iterations of IBP and report the results for IBP(1) and IBP(10).

Algorithm approx-mpe(i) was tested for several values of $i$. Its accuracy increases with increasing $i$, and the algorithm becomes exact for $i > w^*$, where $w^*$ is the induced width of the network.

### 6.1 Exact MPE versus exact belief-update decoding

Before testing the approximation algorithms for bit-wise and block-wise decoding, we tested whether there is a significant difference between the corresponding exact decoding algorithms. We compared exact elim-mpe against exact elim-bel on several types of net-works, including two Hamming code networks, randomly generated networks with different number of parents, and structured code. The results for 1000 input signals, generated randomly for each network, are presented in Table 6.1. When noise is relatively low ($\sigma = 0.3$), both algorithms have practically the same decoding error, while for larger noise ($\sigma = 0.5$) the bit-wise decoding (elim-bel) gives a slightly smaller BER than the block-wise decoding (elim-mpe).

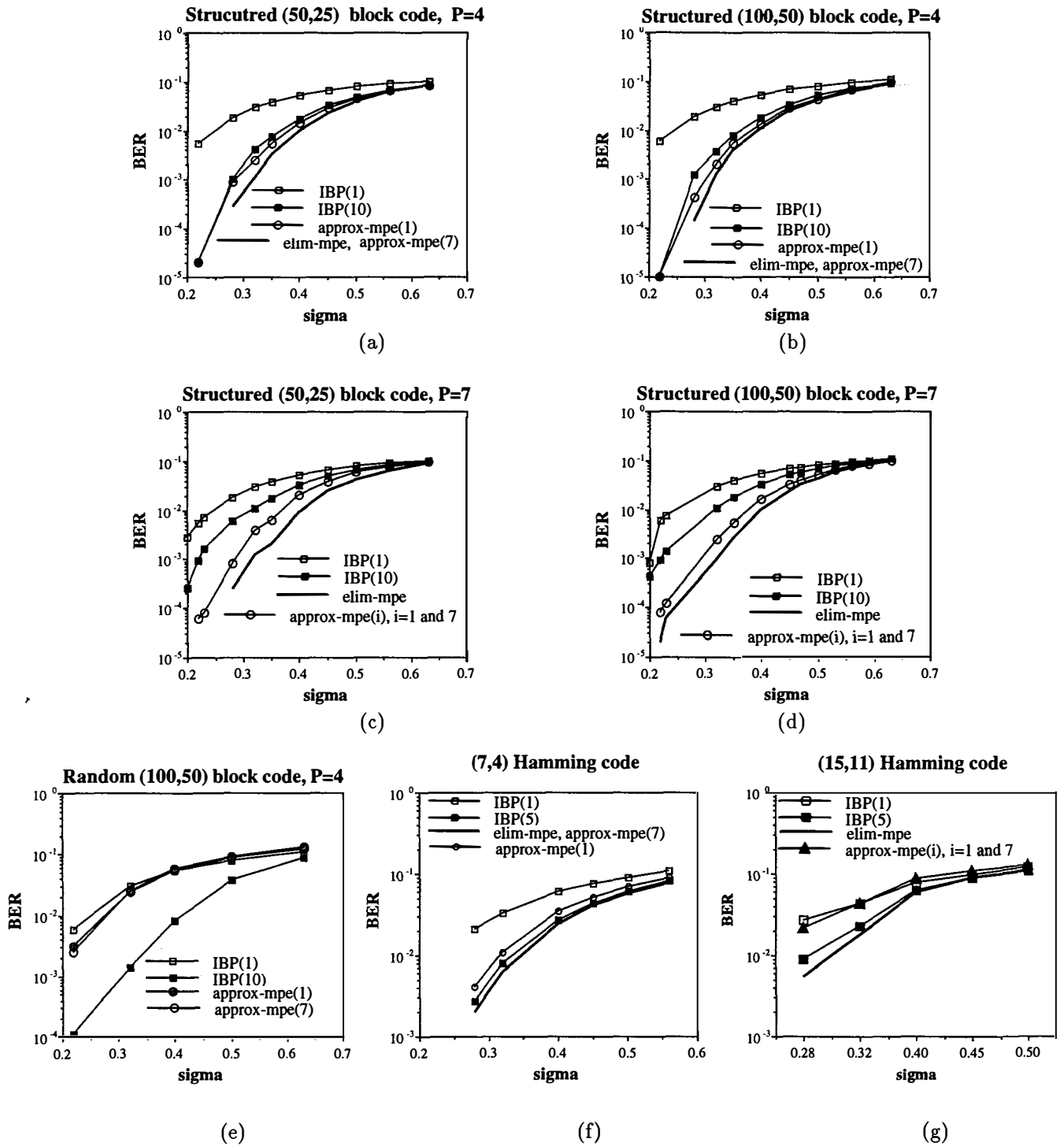

Figure 6: The bit error rate (BER) as a function of the channel noise parameter ($\sigma$) *elim-mpe*, *approx-mpe*(1) and *approx-mpe*(7), and for IBP with 1 and 10 iterations (IBP(1) and IBP(10)) on
1. **Structured block codes** (Figures (a)-(d)) with parameters:
(a) N=50, K=25, P=4; (b) N=100, K=50, P=4; (c) N=50, K=25, P=7; and (d) N=100, K=50, P=7.
2. **Random block code** (Figure (e)) with parameters N=100, K=50, P=4.
3. **Hamming codes** (Figures (f)) with parameters (f) K=4, N=7 and (g) K=11, N=15.
The induced width $w^*$ of the networks was:
(a),(b) $w^* = 6$; (c),(d) $w^* = 12$; (e) $30 \leq w^* \leq 45$; (f) $w^* = 3$; (g) $w^* = 9$.
An average BER is computed on 1000 randomly generated input signals.

### 6.2 Structured parity-check codes

Comparison of the algorithms on the structured parity-check code networks with K=25 and 50, and P=4 and 7 (figure 6(a)-(d)) shows that:
1. as expected, exact elim-mpe decoder always gives the smallest error;
2. IBP(10) is more accurate on average than IBP(1); however, this may not be the case on some particular instances;
3. as expected, approximation algorithm elim-mpe(i) is close to elim-mpe, due to the low induced width of the networks ($w^* = 6$ for P=4, and $w^* = 12$ for P=7).
4. *algorithm elim-mpe(i) outperforms IBP and IBR on all structured networks.*
With increasing the parents set size from P=4 (figures 6(a) and 6(b)) to P=7 (figures 6(c) and 6(d)), the difference between IBP and elim-mpe(i) becomes even more pronounced. On networks with P=7 both approx-mpe(1) and approx-mpe(7) achieve an order of magnitude smaller error than IBP(10).

Next, we consider the results for each algorithm separately, increasing the number of parents from P=4 to P=7 (see also Table 2). We see that the error of IBP(1) practically does not change, the error of the exact elim-mpe changes only slightly, while the error of IBP(10) and approx-mpe(i) increases. However, the BER of IBP(10) increased more dramatically with increased parent set. The induced width of the network was increasing with the increasing parent set size, and affected the quality of both IBP and approx-mpe(i). In case of P=4 (induced width 6), approx-mpe(7) coincides with elim-mpe; in case of P=7 (induced width 12) approximation algorithms do not coincide with elim-mpe. Still, they are better than IBP.

In summary, the results on structured parity-check codes demonstrate that approx-mpe(i), even for i=1, was always at least as accurate as belief propagation with 10 iterations. On networks with larger parent sets approx-mpe(1) and approx-mpe(7) were an order of magnitude better than IBP(10), and about two orders of magnitude better than IBP(1). These results indicate that approx-mpe(i) may be a better decoder than IBP on the networks having relatively small induced width.

### 6.3 Random parity-check code

On randomly generated parity-check networks (Figure 6(e)) the picture was reversed: approx-mpe(i) was worse than IBP(10), although as good as IBP(1). Elim-mpe always ran out of memory on those networks (the induced width exceeded 30). The results are not surprising since approx-mpe(i) is not likely to be accurate if the bound $i$ is much lower than the induced width. However, it is not clear why IBP was much better in this case. We also need to compare our algorithms on other random code generators such as recently introduced in coding community *low-density generator matrix* [3] and *low-density parity-check*[13] code.

### 6.4 Hamming codes

We tested the belief propagation and the mini-bucket approximation algorithms on two Hamming code networks, one with $K = 4, N = 7$, and the other one with $K = 11, N = 15$. The results are shown in figures 6(f) and 6(g). Again, the most accurate results are produced by elim-mpe decoding. Since the induced width of the (7,4) Hamming network is only 3, approx-mpe(7) coincides with the exact algorithm. IBP(1) is much worse than the rest of the algorithms, while IBP(5) is very close to the exact elim-mpe. Algorithm approx-mpe(1) is slightly worse than IBP(5). On the larger Hamming code network, the results are similar, except that both approx-mpe(1) and approx-mpe(7) are inferior to IBP(5), which is inferior to elim-mpe. These results are expected since the induced width of the network is larger ($w^* = 9$), so that approximation with bound 7 or less is suboptimal. Since the networks were quite small, the runtime of all algorithms was less than a second, and the time of IBP(5) was comparable to the time of exact elim-mpe.

In summary, we see that MPE decoding is always superior to IBP decoding. It is expected that an optimal algorithm will be superior, however, since elim-mpe minimizes the block error, while IBP minimizes the bit error, this was not completely clear. For structured parity-check networks having bounded induced width, approx-mpe(i) also outperforms IBP by order(s) of magnitude. However, when the induced width is large (e.g., random parity-check codes), approx-mpe(i) is less accurate than IBP.

## 7 Conclusions

This paper studies empirically the performance of several approximate probabilistic inference algorithms on the probabilistic decoding problem. The mini-bucket approximation algorithms for finding a most probable explanation (MPE) are compared against the following algorithms: exact *elim-mpe, elim-map, elim-bel*, and a variation of Pearl's *belief propagation* algorithm. The latter was shown recently to be a highly effective decoder if applied iteratively to coding networks. We evaluated these algorithms on *linear block codes* of several types: Hamming codes, structured codes with low induced width, and randomly generated codes.

Table 1: BER of exact decoding algorithms *elim-bel* (denoted *bel*) and *elim-mpe* (denoted *mpe*) on several block codes ( average on 1000 randomly generated input signals).

| $\sigma$ | Hamming code | | | | Random code | | | | | | Structured code | |
|---|---|---|---|---|---|---|---|---|---|---|---|---|
| | (7,4) | | (15,11) | | K=10, N=20, P=3 | | K=10, N=20, P=5 | | K=10, N=20, P=7 | | K=25, P=4 | |
| | bel | mpe | bel | mpe | bel | mpe | bel | mpe | bel | mpe | bel | mpe |
| 0.3 | 6.7e-3 | 6.8e-3 | 1.6e-2 | 1.6e-2 | 1.8e-3 | 1.7e-3 | 5.0e-4 | 5.0e-4 | 2.1e-3 | 2.1e-3 | 6.4e-4 | 6.4e-4 |
| 0.5 | 9.8e-2 | 1.0e-1 | 1.5e-1 | 1.6e-1 | 8.2e-2 | 8.5e-2 | 8.0e-2 | 8.1e-2 | 8.7e-2 | 9.1e-2 | 3.9e-2 | 4.1e-2 |

Our results show that in coding networks inducing small width, the mini-bucket algorithms outperformed the iterative belief propagation. However, for networks having a large induced-width (e.g., random codes) and for Hamming codes the iterative approach outperformed the basic mini-bucket algorithm. In all cases, and as is quite expected, decoding using the optimal elim-mpe algorithm was best.

As expected, we observe dependence between the network's induced-width and the quality of the mini-bucket's approach; such dependence is less clear for iterative belief propagation. We also observe increased accuracy as we employ more powerful mini-bucket algorithms.

Our experiments were restricted to networks having small parent sets since the mini-bucket approach and the belief propagation approaches are, in general, time and space exponential in the parent set. This limitation can be eliminated in using the specific structure of coding networks that is a particular (deterministic) case of *causal independence* [12, 21]. Such networks can be transformed into networks having families of size three only. Indeed, in coding practice, the belief propagation algorithm is linear in the family size, thus allowing processing networks of arbitrary family size. We plan to exploit causal independence in the mini-bucket algorithms as well, and hope to extend our experimental evaluation accordingly.

## Acknowledgments

We wish to thank Padhraic Smyth and Robert McEliece for insightful discussions and providing various information about the coding domain.

Table 2: Bit error rate and runtime of the algorithms on structured a block codes. Notation: (1) is IBP(1), (2) is IBP(10), (3) is elim-mpe, (4) is approx-mpe(1), (5) is approx-mpe(7).

| $\sigma$ | BER | | | | | Time | | | | |
|---|---|---|---|---|---|---|---|---|---|---|
| | (1) | (2) | (3) | (4) | (5) | (1) | (2) | (3) | (4) | (5) |
| K=25, P=4, R=1/2, 1000 experiments per row induced width = 6 | | | | | | | | | | |
| 0.28 | 1.8e-2 | 1.0e-3 | 2.8e-4 | 8.8e-4 | 2.8e-4 | 0.02 | 0.16 | 0.03 | 0.01 | 0.04 |
| 0.32 | 3.0e-2 | 4.1e-3 | 1.1e-3 | 2.5e-3 | 1.1e-3 | 0.02 | 0.16 | 0.03 | 0.02 | 0.04 |
| 0.35 | 3.8e-2 | 7.4e-3 | 3.2e-3 | 5.4e-3 | 3.2e-3 | 0.02 | 0.16 | 0.03 | 0.02 | 0.04 |
| 0.40 | 5.2e-2 | 1.7e-2 | 1.0e-2 | 1.4e-2 | 1.0e-2 | 0.02 | 0.16 | 0.03 | 0.02 | 0.04 |
| 0.45 | 6.6e-2 | 3.2e-2 | 2.3e-2 | 2.9e-2 | 2.3e-2 | 0.02 | 0.16 | 0.03 | 0.01 | 0.04 |
| 0.50 | 8.0e-2 | 4.8e-2 | 3.9e-2 | 4.6e-2 | 3.9e-2 | 0.02 | 0.16 | 0.03 | 0.01 | 0.04 |
| 0.56 | 9.0e-2 | 6.4e-2 | 6.2e-2 | 6.4e-2 | 6.2e-2 | 0.02 | 0.16 | 0.03 | 0.01 | 0.04 |
| 0.63 | 1.0e-1 | 8.4e-2 | 8.4e-2 | 8.3e-2 | 8.4e-2 | 0.02 | 0.16 | 0.03 | 0.01 | 0.04 |
| K=25, P=7, R=1/2, 1000 experiments per row induced width = 12 | | | | | | | | | | |
| 0.28 | 1.8e-2 | 6.0e-3 | 2.4e-4 | 8.0e-4 | 8.0e-4 | 0.36 | 3.89 | 1.17 | 0.07 | 0.13 |
| 0.32 | 3.0e-2 | 1.1e-2 | 1.2e-3 | 3.9e-3 | 3.9e-3 | 0.53 | 5.78 | 1.73 | 0.10 | 0.19 |
| 0.35 | 3.8e-2 | 1.7e-2 | 2.0e-3 | 6.1e-3 | 6.1e-3 | 0.33 | 3.55 | 1.07 | 0.06 | 0.12 |
| 0.40 | 5.2e-2 | 3.3e-2 | 8.8e-3 | 2.1e-2 | 2.1e-2 | 0.33 | 3.55 | 1.07 | 0.06 | 0.12 |
| 0.45 | 6.6e-2 | 4.9e-2 | 2.5e-2 | 3.7e-2 | 3.7e-2 | 0.33 | 3.55 | 1.07 | 0.06 | 0.12 |
| 0.50 | 8.0e-2 | 6.5e-2 | 4.2e-2 | 5.9e-2 | 5.9e-2 | 0.33 | 3.55 | 1.07 | 0.06 | 0.12 |
| 0.56 | 9.0e-2 | 8.1e-2 | 6.3e-2 | 7.4e-2 | 7.4e-2 | 0.33 | 3.56 | 1.07 | 0.06 | 0.12 |
| 0.63 | 1.0e-1 | 1.0e-1 | 8.9e-2 | 9.3e-2 | 9.3e-2 | 0.33 | 3.56 | 1.07 | 0.06 | 0.12 |
| K=50, P=4, R=1/2, 1000 experiments per row induced width = 6 | | | | | | | | | | |
| 0.28 | 1.9e-2 | 1.2e-3 | 1.4e-4 | 4.3e-4 | 1.4e-4 | 0.03 | 0.33 | 0.06 | 0.03 | 0.09 |
| 0.32 | 3.0e-2 | 3.7e-3 | 1.3e-3 | 2.0e-3 | 1.3e-3 | 0.03 | 0.33 | 0.06 | 0.03 | 0.09 |
| 0.32 | 3.9e-2 | 7.8e-3 | 3.9e-3 | 5.4e-3 | 3.9e-3 | 0.03 | 0.33 | 0.06 | 0.03 | 0.09 |
| 0.40 | 5.3e-2 | 1.8e-2 | 1.1e-2 | 1.3e-2 | 1.1e-2 | 0.03 | 0.33 | 0.06 | 0.03 | 0.09 |
| 0.45 | 6.8e-2 | 3.3e-2 | 2.5e-2 | 2.8e-2 | 2.5e-2 | 0.03 | 0.33 | 0.06 | 0.03 | 0.09 |
| 0.50 | 8.0e-2 | 5.1e-2 | 4.1e-2 | 4.4e-2 | 4.1e-2 | 0.03 | 0.33 | 0.06 | 0.03 | 0.09 |
| 0.56 | 9.4e-2 | 7.1e-2 | 6.3e-2 | 6.8e-2 | 6.3e-2 | 0.03 | 0.33 | 0.06 | 0.03 | 0.09 |
| 0.63 | 1.1e-1 | 9.2e-2 | 8.9e-2 | 9.5e-2 | 8.9e-2 | 0.03 | 0.33 | 0.06 | 0.03 | 0.09 |
| K=50, P=7, R=1/2, 1000 experiments per row induced width = 12 | | | | | | | | | | |
| 0.32 | 3.0e-2 | 1.1e-2 | 9.6e-4 | 2.4e-3 | 2.4e-3 | 0.66 | 7.15 | 2.82 | 0.12 | 0.24 |
| 0.35 | 3.9e-2 | 1.8e-2 | 2.7e-3 | 5.3e-3 | 5.3e-3 | 0.43 | 4.71 | 1.86 | 0.08 | 0.16 |
| 0.40 | 5.5e-2 | 3.3e-2 | 1.0e-2 | 1.6e-2 | 1.6e-2 | 0.66 | 7.15 | 2.82 | 0.12 | 0.24 |
| 0.45 | 6.8e-2 | 5.1e-2 | 2.4e-2 | 3.2e-2 | 3.2e-2 | 0.43 | 4.71 | 1.86 | 0.08 | 0.16 |
| 0.50 | 8.1e-2 | 6.9e-2 | 4.4e-2 | 5.2e-2 | 5.2e-2 | 0.66 | 7.15 | 2.82 | 0.12 | 0.24 |
| 0.56 | 9.5e-2 | 8.8e-2 | 7.3e-2 | 7.7e-2 | 7.7e-2 | 0.66 | 7.15 | 2.82 | 0.12 | 0.24 |
| 0.59 | 1.0e-1 | 9.6e-2 | 8.3e-2 | 8.8e-2 | 8.8e-2 | 0.69 | 7.47 | 2.94 | 0.13 | 0.25 |
| 0.63 | 1.1e-1 | 1.1e-1 | 9.7e-2 | 1.0e-1 | 1.0e-1 | 0.67 | 7.22 | 2.85 | 0.12 | 0.24 |